\documentclass[10pt,twocolumn,letterpaper]{article}

\usepackage{cvpr}              

\usepackage{graphicx}
\usepackage{amsmath}
\usepackage{amssymb}
\usepackage{booktabs}
\usepackage{epstopdf}
\usepackage[pagebackref=true,breaklinks=true,letterpaper=true,colorlinks,bookmarks=false]{hyperref}
\usepackage{booktabs}
\usepackage{stfloats}
\usepackage{float}
\usepackage{multirow}
\usepackage{bbding}
\usepackage{bm}
\usepackage{amsfonts,amssymb}

\usepackage{caption}
\usepackage[capitalize]{cleveref}
\crefname{section}{Sec.}{Secs.}
\Crefname{section}{Section}{Sections}
\Crefname{table}{Table}{Tables}
\crefname{table}{Tab.}{Tabs.}

\def\cvprPaperID{3029} 
\def\confName{CVPR}
\def\confYear{2022}

\begin{document}

\title{DE-Net: Dynamic Text-guided Image Editing Adversarial Networks}

\author{Ming Tao\textsuperscript{1} \quad 
Bing-Kun Bao\textsuperscript{1} \quad
Hao Tang\textsuperscript{2} \quad
Fei Wu\textsuperscript{1} \quad
Longhui Wei\textsuperscript{3} \quad
Qi Tian\textsuperscript{3} \\
\textsuperscript{1}Nanjing University of Posts and Telecommunications \quad  
\textsuperscript{2}CVL, ETH Zürich \quad 
\textsuperscript{3}Huawei Inc. \\
}

\maketitle

\begin{abstract}

Text-guided image editing models have shown remarkable results. However, there remain two problems.
First, they employ fixed manipulation modules for various editing requirements (e.g., color changing, texture changing, content adding and removing), which results in over-editing or insufficient editing.
Second, they do not clearly distinguish between text-required and text-irrelevant parts, which leads to inaccurate editing.
To solve these limitations, we propose:
(i) a Dynamic Editing Block (DEBlock) which composes different editing modules dynamically for various editing requirements.
(ii) a Composition Predictor (Comp-Pred) which predicts the composition weights for DEBlock according to the inference on target texts and source images.
(iii) a Dynamic text-adaptive Convolution Block (DCBlock) which queries source image features to distinguish text-required parts and text-irrelevant parts.
Extensive experiments demonstrate that our DE-Net achieves excellent performance and manipulates source images more correctly and accurately.
Code is available at \url{https://github.com/tobran/DE-Net}.

\end{abstract}

\section{Introduction}

The last few years have witnessed the great success of Generative Adversarial Networks (GANs) \cite{goodfellow2014generative} for a variety of applications \cite{tang2019multi,zhang2019self,karras2019style,karras2020analyzing,tang2020local}. 
Among them, image editing is one of the important applications of GANs.
Recently, some works tried to bridge the gap between the natural language and image editing \cite{dong2017semantic,liu2020describe,liu2020open,jiang2021language,li2020manigan,li2020lightweight,nam2018text,xu2022predict}. 
Different from other image editing tasks, text-guided image editing aims to translate source images to target images according to given text descriptions.
Due to the advantages of natural language, the text guidance makes the image editing process more controllable and direct.
Moreover, owing to its promising applications, this task has attracted many researchers and achieved many significant progresses~\cite{dong2017semantic,nam2018text,li2020manigan,li2020lightweight}. 

\begin{figure}[t] \small
  \centering
  \includegraphics[width=\linewidth]{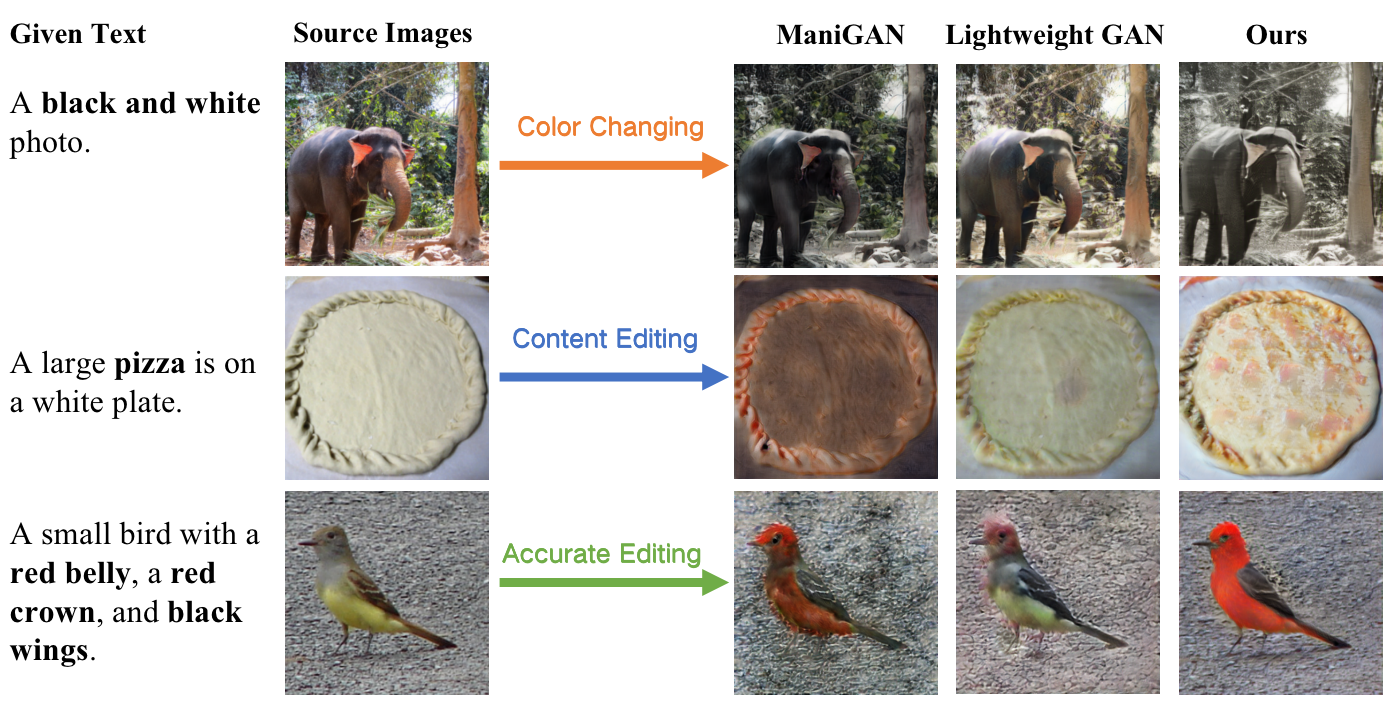}
  \vspace{0.1cm}
  \caption{Benefiting from dynamic editing design, our DE-Net can deal with various editing tasks (e.g., color changing, content editing). Furthermore, the text-adaptive convolution in DCBlock enables more accurate manipulations.}
  \label{fig0}
\end{figure}

However, current text-guided image editing models have two problems.
First, they tackle various editing requirements equally with fixed manipulation modules.
Different from other image-to-image tasks (e.g., image colorization, style transfer, and image inpainting), which design effective models for specific tasks, text-guided image editing is an open-target task.
It contains various editing tasks (e.g., color changing, texture changing, content adding, and removing) prompted by various text guidance.
But current models ignore the difference between varying editing requirements and employ fixed manipulation modules for all editing tasks.
It leads the whole network tends to make a trade-off for different editing tasks.
These compromises limit the manipulation ability and result in over-editing or insufficient editing.
As the result shown in Figure~\ref{fig0}, previous models \cite{li2020manigan,li2020lightweight} fail to deal with different editing requirements.

Second, the editing preciseness is deficient, as current manipulation modules cannot clearly distinguish between text-required and text-irrelevant parts.
Text-visual feature concatenations \cite{dong2017semantic,nam2018text} and ACM module \cite{li2020manigan,li2020lightweight} are widely used to manipulate image features.
However, naive concatenation treats text-required and text-irrelevant parts equally, resulting in text information overwriting the source image features.
And the ACM module predicts editing parameters from the encoded source images directly, without introducing text information, which makes it unable to accurately edit text-required parts and preserve text-irrelevant parts.
As the result shown in Figure~\ref{fig0}, the previous model \cite{li2020manigan,li2020lightweight} diffuses the text information to the whole image.

To address these problems, we propose a novel Dynamic text-guided image Editing adversarial Network (i.e., DE-Net). 
Different from previous works \cite{dong2017semantic,nam2018text,li2020manigan,li2020lightweight} which employ fixed editing blocks, our DE-Net employs Dynamic Editing Block (DEBlock).
The DEBlock combines spatial- and channel-wise manipulations dynamically for different editing requirements.
Furthermore, we propose the Composition Predictor (Comp-Pred) to predict the composition weights of DEBlock.
The DEBlock and Comp-Pred enable our network can deal with various editing tasks through suitable combinations of spatial- and channel-wise manipulations.
In addition, to improve the editing preciseness of the manipulation module, we propose a novel Dynamic text-adaptive Convolution Block (DCBlock).
The DCBlock employs a dynamic text-adaptive convolutional layer that can automatically adjust the kernel weights according to the given text guidance. This enables our model to distinguish text-required and text-irrelevant image features through querying the visual features according to text guidance.
Moreover, we add a lightweight semantic decoder that provides global-local visual features for DCBlock to obtain a more accurate prediction of editing parameters.

Overall, our contributions can be summarized as follows:
\begin{itemize}
    \item We propose a novel Dynamic text-guided Editing Block (DEBlock) to enable our model can deal with a variety of editing tasks adaptively through the dynamic composition of different editing modules.
    \item We propose a novel Composition Predictor (Comp-Pred), which predicts the composition weights for DEBlock according to the inference on text and visual features.
    \item We propose a novel Dynamic text-adaptive Convolution Block (DCBlock) which can distinguish text-required parts and text-irrelevant parts of source images.
    \item Compared with the current state-of-the-art methods, our DE-Net achieves much better performance on commonly used public datasets.
\end{itemize}

\section{Related Work}

\noindent \textbf{Text-to-Image Synthesis} bridges the text information and image generation.
Based on Generative Adversarial Network (GAN) \cite{goodfellow2014generative}, some researchers tried to bridge the text and image modality.
\cite{reed2016generative} first employed the conditional Generative Adversarial Network (cGAN) to synthesize plausible images from given text descriptions.
\cite{zhang2017stackgan,zhang2018stackgan} proposed the stacked architecture, which stacks multiple generators and discriminators to synthesize high-resolution images. 
\cite{xu2018attngan} introduced the attention mechanism to fuse the word information and image features.
This text-image attention mechanism has been widely applied in many text-related generative works \cite{zhu2019dm,cheng2020rifegan}.
\cite{tao2022df} proposed the DF-GAN with deep fusion block and matching-aware gradient penalty, which enables one-stage high-quality generation.
Recently, some large transformer-based text-to-image methods~\cite{ramesh2021zero,lin2021m6,ding2021cogview} show excellent performance on complex image synthesis.
They tokenize the images and take the image tokens and word tokens to make auto-regressive training by a unidirectional transformer \cite{radford2019language,brown2020language}.

\noindent \textbf{Text-Guided Image Editing} aims to manipulate source images according to given text descriptions.
For example, \cite{dong2017semantic} first proposed SISGAN based on an encoder-decoder architecture. 
It concatenates the encoded representations with text semantics to manipulate source images.
The TAGAN \cite{nam2018text} introduced a word-level attention mechanism in a discriminator to classify fine-grained attributes independently.
However, both SISGAN \cite{dong2017semantic} and TAGAN \cite{nam2018text} only preserve coarse visual features from source images.
\cite{li2020manigan} proposed the ManiGAN, which is a two-stage architecture with the ACM and DCM modules.
The ACM module synthesizes text-matching images while preserving a rough shape from source images.
And the DCM module rectifies mismatched image features and completes missing details.
Then, \cite{li2020lightweight} proposed a more efficient word-level discriminator to facilitate training a lightweight generator.
Recently, the ManiTrans \cite{wang2022manitrans} employ the extra knowledge of pre-training autoregressive transformer for text-guided image editing.
These methods have shown significant progress in text-guided image editing tasks.
Recently, some text-guided image editing works focus on face editing \cite{xia2021tedigan,xia2021towards,patashnik2021styleclip}. 
They employ a pretrained StyleGAN \cite{karras2019style} and manipulate its latent according to face descriptions.

In this work, we focus on the main research direction of the text-guided image editing task, which deals with open-target image editing \cite{dong2017semantic,nam2018text,li2020manigan,li2020lightweight}. 
Unlike previous models, our proposed DE-Net employs a Dynamic Editing Block (DEBlock) with a Composition Predictor (Comp-Pred) to combine the editing processes adaptively for various editing requirements.
Moreover, we equip our DE-Net with a Dynamic text-adaptive Convolution Block (DCBlock) which can distinguish between text-required and text-irrelevant image features.
Compared with previous works, our model edits the text-required image features more efficiently and preserves text-irrelevant source image features more accurately.

\section{The Proposed DE-Net}

\begin{figure*}[t] \small
  \centering
  \includegraphics[width=\linewidth]{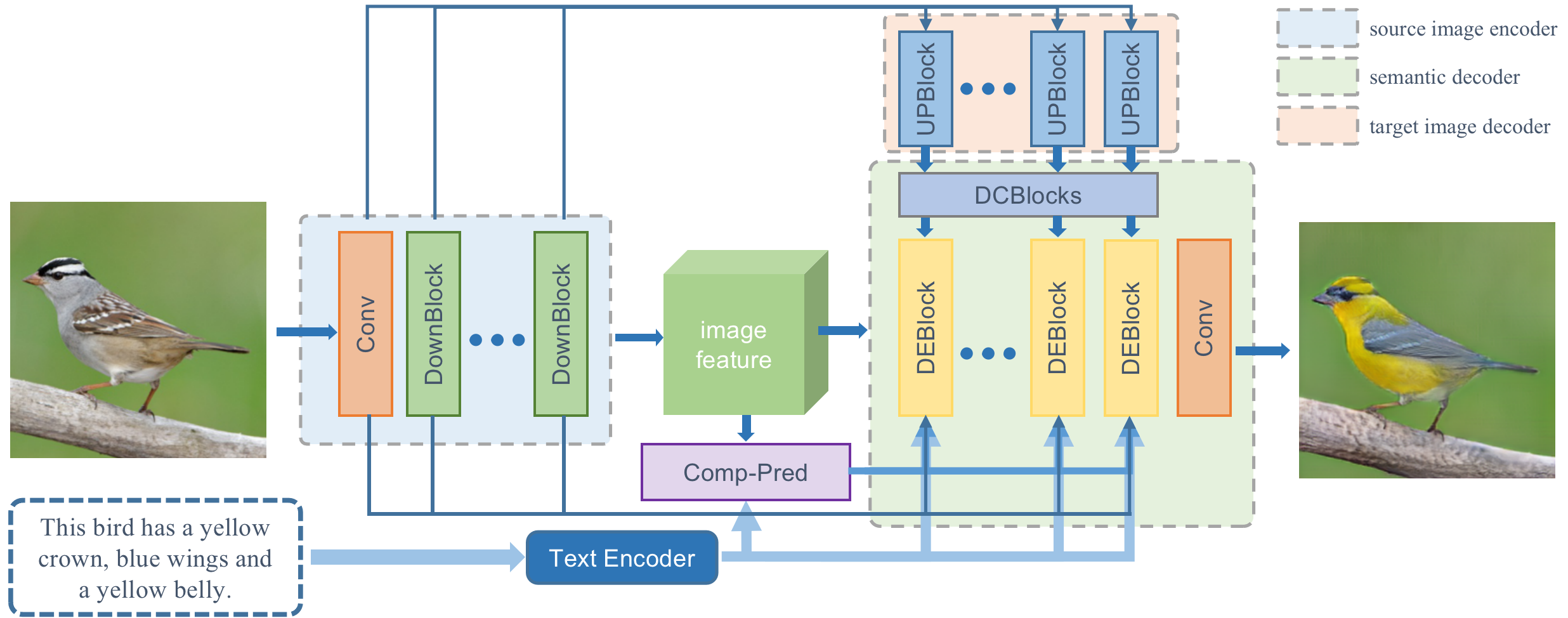}
  \vspace{0.2cm}
  \caption{The architecture of the proposed DE-Net. The DE-Net comprises a source image encoder, a target image decoder, a semantic decoder, a Composition Predictor (Comp-Pred), and a pretrained text encoder \cite{xu2018attngan}. The DEBlock and DCBlock are introduced in the target image decoder to enable effective and accurate manipulations, respectively.}
  \label{fig2}
\end{figure*}

In this paper, we propose a novel GAN model for text-guided image editing named Dynamic text-guided Image Editing adversarial Networks (DE-Net) (see Figure~\ref{fig2}). 
To manipulate source images according to text guidance more effectively and accurately, we propose: 
(i) a novel manipulation module called Dynamic Editing Block (DEBlock), which manipulates the image features adaptively through the dynamic combination between spatial and channel-wise editing.
(ii) a novel Composition Predictor (Comp-Pred) that can predict the combination weights according to the inference on text and visual features.
(iii) a novel Dynamic text-adaptive Convolution Block (DCBlock) that improves the editing preciseness through querying the visual features according to text guidance.
In the following of this section, we first present the overall structure of our DE-Net and then introduce the Comp-Pred, DEBlock, and DCBlock in detail.

\subsection{Dynamic Text-guided Image Editing Framework}

Different from previous models \cite{dong2017semantic,nam2018text,li2020manigan,li2020lightweight,wang2022manitrans}, our DE-Net is based on a novel dynamic text-guided image editing framework. 
First, it can compose suitable editing processes based on input images and texts dynamically.
Second, it can distinguish between text-required and text-irrelevant image features through dynamic text-adaptive convolution.

As shown in Figure~\ref{fig2}, the whole editing framework is composed of a source image encoder, a target image decoder, a Composition Predictor (Comp-Pred), a semantic decoder, and a pre-trained text encoder~\cite{li2020manigan,li2020lightweight}.
To introduce multi-scale source visual features for manipulation and preservation, we add full skip connections between the source image encoder and two decoders.
In DE-Net, the source image is first encoded to $4{\times}4$ resolution by the source image encoder.
Then, the encoded source image feature is fed into the semantic decoder and target image decoder, respectively.
The semantic decoder comprises several upsampling and convolutional layers (UPBlock).
It can capture long-range visual information and decode the global-local semantic features for the DCBlock through the growing size of semantic features.
The target image decoder is composed of multiple DEBlocks and DCBlocks.
The DEBlock manipulates source image features both from spatial and channel dimensions and dynamically combines these two manipulations at different image scales.
And the Comp-Pred predicts the combination weights for DEBlock according to the inference on text and visual features.
The DCBlock adjusts its candidate convolution kernels and applies text-adaptive convolution on global-local semantic features provided by the semantic decoder.
It enables our DE-Net to distinguish between text-required and text-irrelevant parts.

To stabilize the training process of adversarial learning, we further introduce the target-aware discriminator \cite{tao2022df} in DE-Net.
Finally, our DE-Net can be formulated as:
\begin{equation}
 \begin{split}
 L_D = &-\mathbb{E}[min(0,-1+D(x,e))]\\
       &-(1/2)\mathbb{E}[min(0,-1-D(G(x,\hat{e}),\hat{e}))]\\
       &-(1/2)\mathbb{E}[min(0,-1-D(x,\hat{e}))]\\
       &+k\mathbb{E}[(\|\nabla_{x}D(x,e)\|+\|\nabla_{e}D(x,e)\|)^{p}], \\
 L_G = &-\mathbb{E}[D(G(x,\hat{e}),\hat{e})]+\lambda_1\mathbb{E}[\|G(x,\hat{e})-x\|_2]\\
       &-\lambda_2\mathbb{E}[S(G(x,\hat{e}),\hat{e})], \\
 \end{split} 
\end{equation}
where $x$ denotes the source images; $e$ denotes the source sentence embeddings, $\hat{e}$ denotes the target sentence embeddings, $G$ denotes the generator network, $D$ denotes the discriminator network, $k$ and $p$ are two hyperparameters of discriminator to balance the effectiveness of gradient penalty.
$S$ represents the cosine similarity between the encoded visual and text features predicted by pre-trained DAMSM network \cite{xu2018attngan,li2020manigan,li2020lightweight}.
$\lambda_1$ and $\lambda_2$ are two balancing coefficients of the generator.

\subsection{Composition Predictor}

\begin{figure*}[t] \small
  \centering
  \includegraphics[width=\linewidth]{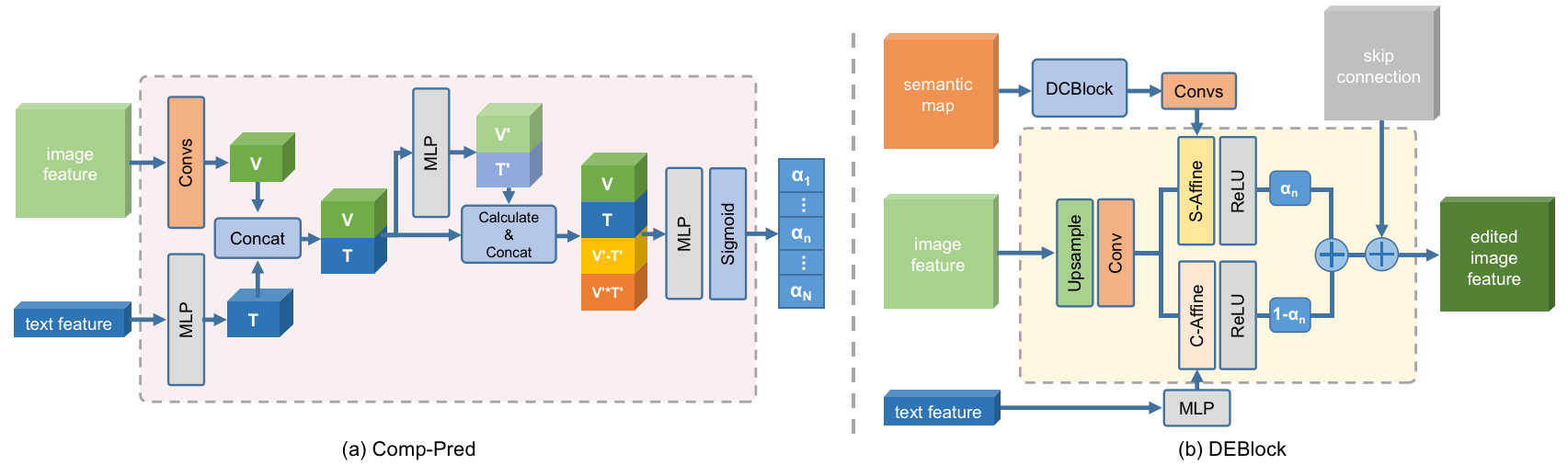}
  \caption{Illustration of the proposed Comp-Pred and DCBlock. (a) The Comp-Pred predicts the combination weights for DEBlock. (b) The DEBlock composes suitable editing processes for each input pair according to the combined weights of C-Affine and S-Affine.}
  \label{fig3}
\end{figure*}

To enable our DE-Net to compose suitable editing processes based on current source images and text guidance, we propose the Composition Predictor (Comp-Pred).
As shown in Figure~\ref{fig2} and Figure~\ref{fig3}, for visual information, it takes the encoded $4{\times}4$ image feature and maps it to the visual vector $\bm{V}$ through two convolution layers.
For text information, it takes the sentence embedding provided by the text encoder, and maps it to a text vector $\bm{T}$ through one-hidden-layer MLP.
Then we concatenate the $\bm{V}$ and $\bm{T}$ and get the $\bm{V'}$ and $\bm{T'}$ through an MLP to close the gap between the visual and text domain.
The difference $\bm{V'}-\bm{T'}$ and the element-wise product $\bm{V'}*\bm{T'}$ are calculated and concatenated with the original vectors $\bm{V}$ and $\bm{T}$ to enhance the distance between text and visual information.
Lastly, we adopt an MLP and a Sigmoid function to predict the combination weights $\bm{\alpha}$ of spatial-wise and channel-wise editing for $N$ DEBlocks in the target image decoder.
The $\bm{\alpha}_{n}$ is a vector and its length is the channel size of the visual feature after convolution layer in each DEBlock.

\subsection{Dynamic Text-guided Image Editing Block}

As mentioned before, the existing naive concatenation \cite{dong2017semantic,nam2018text} or ACM module \cite{nam2018text,li2020lightweight} cannot make an efficient interaction between text guidance and source image features.
Moreover, these manipulation modules treat various editing requirements equally, ignoring the different model requirements.
To address these problems, we first decompose the visual feature editing process into spatial and channel-wise editing.
Then, we propose a novel DEBlock to re-compose suitable editing processes for each input pair through the dynamic combination of spatial and channel-wise affine transformation.

As shown in Figure~\ref{fig3}(a), the DEBlock is composed of upsampling layer, convolution layer, and two kinds of affine layers.
One is Channel-wise Affine layer (C-Affine), the other is Spatial Affine layer (S-Affine).
To fully manipulate source image features, our DEBlock manipulates both spatial and channel dimensions and combines these two manipulations dynamically through combination weights $\bm{\alpha}_{n}$ predicted by Comp-Pred.
Specifically, in C-Affine, the text sentence vector $\bm{t}$ is sent to two different one-hidden-layer MLPs to predict the scaling parameters $\bm{\gamma}_{c}$ and shifting parameters $\bm{\theta}_{c}$ for each channel:
\begin{equation}
\bm{\gamma}_{c} = {\rm MLP}_1(\bm{t}),\qquad\bm{\theta}_{c} = {\rm MLP}_2(\bm{t}).
\end{equation}
After obtaining $\bm{\gamma}_{c}$ and $\bm{\theta}_{c}$ parameters from MLPs, the C-Affine transformation can be formally expressed as follows:
\begin{equation}
{\rm C{-}Aff}(\bm{x}_{c,h,w}|\bm{t})=\bm{\gamma}_{c}\cdot{\bm{x}_{c,h,w}}+\bm{\theta}_{c}.
\end{equation}
where $\bm{x}$ is the image feature; $c$, $h$, and $w$ denotes the channel size, feature height, and feature width, respectively.

Unlike the C-Affine, the S-Affine applies affine transformations on the spatial dimension of image features. 
The spatial scaling parameters $\bm{\gamma}_{h,w}$ and shifting parameters $\bm{\theta}_{h,w}$ are predicted by the DCBlock and two convolution layers for each pixel in image feature.
As shown in Figure~\ref{fig4}, the DCBlock employs the text-adaptive convolution layer to get the text-attended visual features $\bm{f}$.
Then, the spatial scaling and shifting parameters are predicted by two convolution layers on the predicted text-attended visual features:
\begin{equation}
\bm{\gamma}_{h,w} = {\rm Conv}_1(\bm{f}),\qquad\bm{\theta}_{h,w} = {\rm Conv}_2(\bm{f}).
\end{equation}
Thus, the S-Affine can be formally expressed as follows:
\begin{equation}
{\rm S{-}Aff}(\bm{x}_{c,h,w}|\bm{t})=\bm{\gamma}_{h,w}\cdot{\bm{x}_{c,h,w}}+\bm{\theta}_{h,w}.
\end{equation}

To fully manipulate source image features, our model integrates the C-Affine and S-Affine in DEBlock.
And the $n$th DEBlock in the target image decoder combines these two manipulations dynamically through the combination weights $\bm{\alpha}_{n}$ predicted by Comp-Pred:
\begin{equation}
 \begin{aligned}
 {\rm{DEBLK}}(\bm{x}|\bm{t}) = &{\rm C{-}Aff}(\bm{x}|\bm{t})\cdot{\bm{\alpha}_{n}}\\
                                &+ {\rm S{-}Aff}(\bm{x}|\bm{f})\cdot({1-\bm{\alpha}_{n}}).
 \end{aligned} 
\end{equation}

Compared with previous manipulation modules, our DEBlock fully manipulates source image features both from spatial and channel dimensions and considers different contributions of these two dimensions at different image scales.
Moreover, our experiments and ablation studies also demonstrate the effectiveness of the DEBlock.

\subsection{Dynamic Text-adaptive Convolution Block}

\begin{figure}[t] \small
  \centering
  \includegraphics[width=\linewidth]{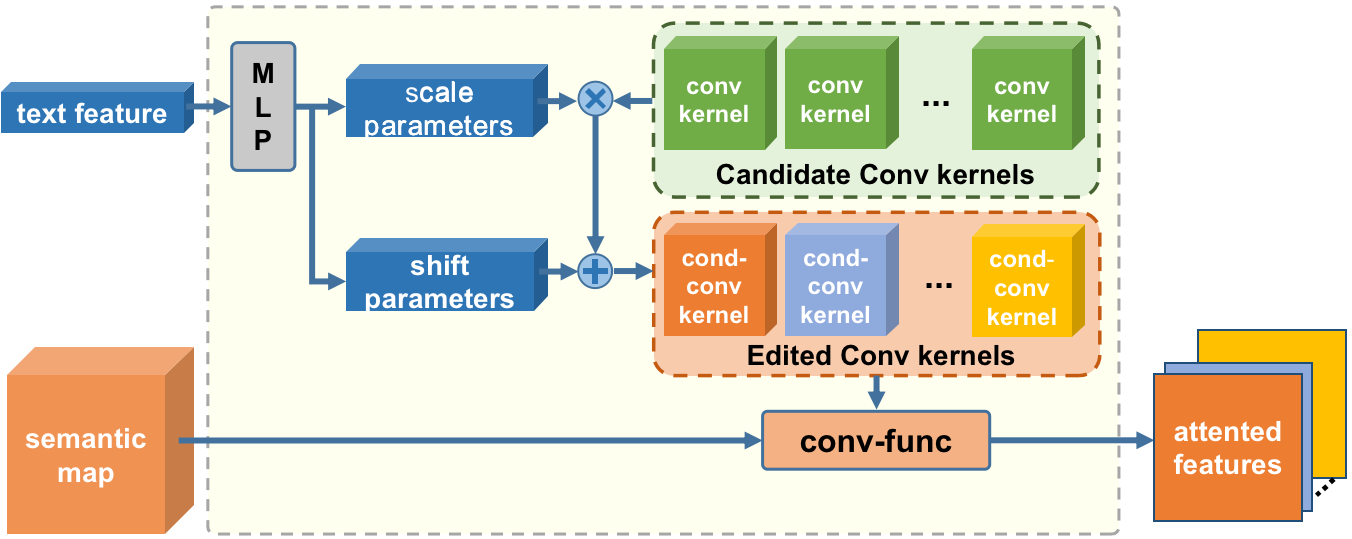}
  \vspace{0.1cm}
  \caption{Illustration of the DCBlock. The text-attended features are predicted by the text-conditioned convolution layer.}
  \label{fig4}
\end{figure}

To improve the editing preciseness of the manipulation module, we propose the DCBlock as an editing positioning module.
As shown in Figure~\ref{fig4}, the weights of convolution kernels in DCBlock are adjusted according to the given text guidance.
However, directly generating all parameters of the convolution kernel not only requires a high computational cost but also results in overfitting to train text descriptions. 
To address this problem, we apply scaling and shifting operations for each candidate convolution kernel.
First, we employ two MLPs to predict the scaling parameters $\bm{\varphi}_{c\_in}$ and shifting parameters $\bm{\omega}_{c\_in}$ for the convolution kernel.
The modulation on one convolution kernel can be formally expressed as follows:
\begin{equation}
{\rm Mod}(\bm{k})=\bm{\varphi}_{c\_in}\cdot{\bm{k}_{c\_in,k\_h,k\_w}}+\bm{\omega}_{c\_in},
\end{equation}
where $\bm{k}_{c\_in,k\_h,k\_w}$ is the convolution kernel; $c\_in$, $k\_h$, and $k\_w$ denotes the input channel size, kernel height, and kernel width, respectively.

With this text-guided adjustment for candidate convolution kernels, the dynamic convolutional (Dy-Conv) operation acts as a querying process according to text descriptions.
However, as the semantic information in one convolution kernel is limited, it is hard to incorporate the whole text semantic information into one convolution kernel.
Inspired by the multi-head attention mechanism in Transformer \cite{vaswani2017attention}, we employ multiple dynamic convolution kernels as different attention heads.
In our work, we set 8 dynamic convolution kernels in DCBlock.
It allows our model to jointly attend to different image features from different text representation subspaces.
As shown in Figure~\ref{fig4}, each dynamic convolution kernel predicts an attended feature map according to partial text semantics.
Then, we stack these attention features along the channel direction as a whole attended feature map for predicting the spatial affine parameters for DEBlock.

However, directly applying Dy-Conv operations on image features from the source image encoder is inefficient.
Each pixel of image features in the source image encoder only contains the semantic information of its local neighborhood through downsampling layers.
It cannot capture long-range contextual information, resulting in low accuracy and efficiency of Dy-Conv.
To cope with this problem, we employ a semantic decoder to decode global-local semantics, as shown in Figure~\ref{fig2}.
The source image is first downsampled through the image encoder and then upsampled by the semantic decoder with full skip connections from the source image encoder.
Through this encoder-decoder structure with full skip-connections, the semantic decoder can capture long-range visual information, and each pixel of decoded semantic features contains global-local semantic information.
We apply Dy-Conv operations on the global-local semantic features to achieve more accurate queried results.
After that, we employ two convolutional layers to predict the spatial scaling and shifting parameters for DEBlock.

The proposed DCBlock enables our DE-Net to distinguish between text-required and text-irrelevant parts.
Furthermore, the semantic decoder provides the global-local semantic features, and the multi-head dynamic convolutional operations query the global-local semantic features from different text representation subspaces.
Armed with the DCBlock, the target image decoder can manipulate source image features more accurately.

\section{Experiments}

\begin{figure*}[t] \small
  \centering
  \includegraphics[width=\linewidth]{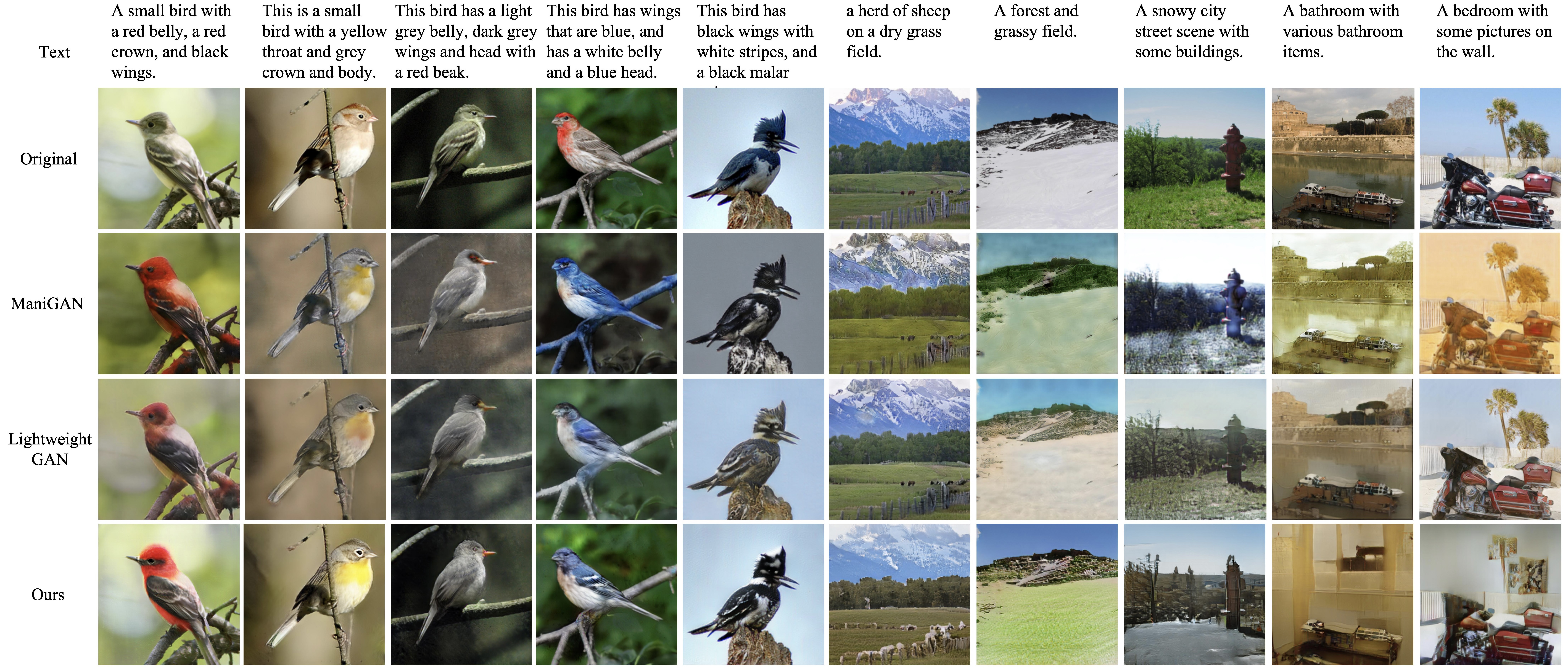}
  \caption{Qualitative comparison between different methods on the test set of CUB and COCO.}
  \label{fig5}
\end{figure*}

In this section, we introduce the datasets, training details, and evaluation metrics used in our experiments. Then we compare the text-guided image editing performance with previous models quantitatively and qualitatively. 

\noindent{\bf Datasets.} We conduct experiments on two challenging datasets: CUB bird \cite{wah2011caltech} and COCO \cite{lin2014microsoft}.
For the CUB bird dataset, there are 11,788 images belonging to 200 bird species, with each image corresponding to ten language descriptions.
For the COCO dataset, it contains 80k images for training and 40k images for testing. 
Each image corresponds to 5 language descriptions.

\noindent{\bf Training and Evaluation Details.} 
We employ the Adam optimizer \cite{kingma2014adam} with $\beta_{1}{=}0.0$ and $\beta_{2}{=}0.9$ to train our model. 
According to the Two Timescale Update Rule (TTUR) \cite{heusel2017gans}, the learning rate is set to 0.0001 for the generator and 0.0004 for the discriminator. 
The hyper-parameters of the discriminator $k$ and $p$ are set to 2 and 6 as \cite{tao2022df}.
The hyper-parameters of the generator $\lambda_1$ and $\lambda_2$ are set to 40 and 4 for all the datasets.

Following previous works \cite{wang2022manitrans, li2020manigan}, we employ the Inception Score (IS) \cite{salimans2016improved}, CLIP-score, L2-error, and Manipulation Precision (MP) to evaluate the performance. 
Higher IS means higher fidelity of the edited images. 
CLIP-score is the text-image cosine similarity calculated by a pretrained CLIP network~\cite{xu2018attngan} to measure the quality of text-guided manipulation.
L2-error is the $L_{2}$ pixel difference between edited images and source images to measure the quality of source feature preserving.
Lastly, MP simultaneously measures the quality of text-guided manipulation and source feature preserving.
It can be formulated as:
\begin{equation}
{\rm MP}=(1-{\rm L2\_error})\times{\rm (CLIP\_score)}.
\end{equation}

It must be pointed out that the Fr\'echet Inception Distance (FID) \cite{heusel2017gans} widely adopted in text-to-image synthesis is not suitable for text-guided image editing.
The FID calculates the Fr\'echet distance between synthetic images and ground truth in the feature space of a pre-trained Inception v3 network.
But there is no ground truth (human edited images) in this task, and calculating with source images makes the FID prefer lazy models which do not edit input images.
So we adopt the IS to evaluate the image fidelity as previous works \cite{wang2022manitrans, li2020manigan}.

\begin{table*}[t] \small
\centering

\resizebox{0.95\linewidth}{!}{%
\begin{tabular}{l|c|c|c|c|c|c|c|c}
\toprule
\multirow{2}*{Model}   & \multicolumn{4}{c|}{CUB}                                                              & \multicolumn{4}{c}{COCO}       \\ 
\cline{2-9} 
                       & IS $\uparrow$  & CLIP-score $\uparrow$   & L2-error $\downarrow$   & MP $\uparrow$    & IS $\uparrow$   & CLIP-score $\uparrow$  & L2-error $\downarrow$   & MP $\uparrow$   \\ \midrule
TAGAN                  & 3.72           & 0.202                   & 0.185                   & 0.164            & 9.33            & 0.125                  & 0.102                   & 0.112           \\
ManiGAN                & 4.19           & 0.213                   & 0.051                   & 0.202            & 22.60           & 0.119                  & 0.031                   & 0.114           \\
DF-GAN                 & 4.54           & 0.221                   & 0.046                   & 0.210            & 20.10           & 0.129                  & 0.027                   & 0.125           \\
Lightweight-GAN        & 4.66           & 0.188                   & 0.132                   & 0.163            & 24.80           & 0.136                  & 0.025                   & 0.132           \\
ManiTrans              & 5.02           & 0.235                   & 0.013                   & 0.231            & 21.45           & 0.131                  & 0.017                   & 0.128           \\
DE-Net (Ours)          & \textbf{5.08}  & \textbf{0.240}          & \textbf{0.010}          & \textbf{0.237}   & \textbf{25.81}  & \textbf{0.192}         & \textbf{0.015}          & \textbf{0.189}  \\ \bottomrule
\end{tabular}}
\caption{Quantitative comparison between different methods on the test set of CUB and COCO.}
\label{table1}
\end{table*}

\subsection{Quantitative Comparisons}

We compare the proposed method with several state-of-the-art text-guided image editing models, including TAGAN \cite{nam2018text}, ManiGAN \cite{li2020manigan}, ManiTrans \cite{wang2022manitrans}.
We also modify the state-of-the-art text-to-image model DF-GAN \cite{tao2022df} and compare our model with it.
As shown in Table~\ref{table1}, our proposed DE-Net achieves the highest IS, CLIP-score, MP, and the lowest L2-error on the CUB and COCO test dataset.
The highest IS shows that our method can produce more realistic manipulated results.
And the highest CLIP-score, MP, and lowest L2-error prove that our method can manipulate text-required image parts and preserve text-irrelevant parts more correctly and accurately.
The extensive quantitative evaluation results on CUB and COCO demonstrate the superiority and effectiveness of our proposed DE-Net.
Furthermore, the advantages of our model are more obvious in the complex COCO images.
It demonstrates the significant improvements of our DE-Net when dealing with various editing requirements, e.g., textual changing, background changing, object adding and removing.

\subsection{Qualitative Comparisons}

In this subsection, we compare the images synthesized by ManiGAN \cite{li2020manigan}, Lightweight GAN \cite{li2020lightweight}, and our DE-Net. 
We evaluate the quality of the edited images from two aspects, text-related source feature manipulation and text-irrelevant source feature preservation.
First, we compare the quality of the edited images on the CUB dataset, then compare the results on the more challenging COCO dataset.

As the results shown in 1$^{st}$, 2$^{nd}$, and 4$^{th}$ columns, both ManiGAN \cite{li2020manigan} and Lightweight GAN \cite{li2020lightweight} cannot accurately synthesize the ``black wings'', ``yellow throat'', and ``white belly'', respectively.
Moreover, they tend to change the color of twigs and backgrounds (see 1$^{st}$, 3$^{rd}$, 4$^{th}$, and 5$^{th}$ columns).
While our DE-Net can manipulate these features successfully and preserve the text-irrelevant image features like twigs and backgrounds more accurately. 
The superiority is more obvious on the challenging COCO dataset, which contains various editing tasks. 
As the results shown in the 6$^{th}$, 7$^{th}$, and 10$^{th}$ columns, both ManiGAN \cite{li2020manigan} and Lightweight GAN \cite{li2020lightweight} cannot synthesize the ``a herd of sheep'', ``grassy field'', and ``some pictures'', respectively.
These results show that they only tend to edit the colors of source images (see 8$^{th}$, 9$^{th}$, and 10$^{th}$ columns).
However, our DE-Net can both edit the color (see 6$^{th}$, and 7$^{th}$ columns), texture (see 7$^{nd}$ and 8$^{th}$ columns) of source images, and synthesize new contents in target images (see 6$^{th}$ and 10$^{th}$ columns).
Moreover, both ManiGAN \cite{li2020manigan} and Lightweight GAN \cite{li2020lightweight} tend to edit the text-irrelevant regions.
For example, the color of the sky is also painted with the color of ``grassy field'' (see 7$^{th}$ column). 
While our DE-Net can preserve these text-irrelevant features or manipulate them to adapt to the text description (see 6$^{th}$, and 7$^{th}$ columns).

The qualitative comparison on both CUB and COCO datasets shows that our DE-Net can manipulate the source images according to the text guidance while preserving text-irrelevant image features more effectively and accurately.

\subsection{Ablation Study}

To verify the effectiveness of different components in DE-Net, we conduct ablation studies on the COCO test dataset.
The components include Dynamic Text-guided Image Editing Block (DE), Composition Predictor (CP), Dynamic Text-adaptive Convolution Block (DC), and the Semantic Decoder (SD). 
The results are provided in Table~\ref{table2}.

\noindent \textbf{Baseline. } Our baseline is based on an encoder-decoder architecture that employs the target-way discriminator \cite{tao2022df}.
The baseline concatenates the text features with the encoded image features. 

\noindent \textbf{Effect of DEBlock (DE). } The experimental results show that the DEBlock achieves better performances than only employing channel-wise (B w/CAFF) or spatial-wise affine transformation (B w/SAFF).
The results demonstrate that our Dynamic Editing strategy, which composes suitable editing processes dynamically based on input text-image pairs, is more effective than fixed editing modules.

\noindent \textbf{Effect of Comp-Pred (CP). } Removing the Comp-Pred (B w/ DE w/o CP) reduces the editing performance.
To further evaluate the effectiveness of the difference and element-wise product in Comp-Pred, we remove them from CP and name the new module CP*. 
Compared with CP*, the Comp-Pred (B w/ DE w/ CP) achieves better performances.
The comparison results prove the effectiveness of Comp-Pred.

\begin{table}[t] \small
\centering

  	\resizebox{0.95\linewidth}{!}{%
\begin{tabular}{l|c|c|c|c}
\toprule
Method                           & IS $\uparrow$   & CLIP-score $\uparrow$  & L2 $\downarrow$  & MP $\uparrow$      \\ \midrule
Baseline (B)                     & 11.08            & 0.115              & 0.057            & 0.108              \\ \hline
B w/ CAFF                        & 12.16            & 0.124              & 0.051            & 0.117              \\
B w/ SAFF                        & 13.33            & 0.131              & 0.047            & 0.124              \\
B w/ DE w/ CP                    & 17.21            & 0.155              & 0.031            & 0.150              \\ \hline
B w/ DE w/o CP                   & 15.36            & 0.141              & 0.040            & 0.135              \\  
B w/ DE w/ CP*                   & 14.13            & 0.145              & 0.039            & 0.139              \\ \hline
B w/ DE w/ CP w/OC               & 18.47            & 0.160              & 0.030            & 0.155              \\
B w/ DE w/ CP w/DC1              & 20.21            & 0.171              & 0.025            & 0.166              \\
B w/ DE w/ CP w/DC4              & 22.18            & 0.176              & 0.021            & 0.172              \\
B w/ DE w/ CP w/DC16             & 21.01            & 0.179              & 0.021            & 0.175              \\
B w/ DE w/ CP w/DC               & 23.11            & 0.180              & 0.018            & 0.176              \\ \hline
B w/ DE w/ CP w/DC w/SD (DE-Net)          & \textbf{25.81}  & \textbf{0.192}     & \textbf{0.015}   & \textbf{0.189}     \\ \bottomrule
\end{tabular}}
\caption{Ablation Study of DE-Net on the test set of COCO.}
\label{table2}
\vspace{-0.4cm}
\end{table}

\noindent \textbf{Effect of DCBlock (DC). } To verify the effectiveness of DCBlock, we replace it with an Ordinary Convolution layer (OC).
The results show that the DCBlock enables our model to manipulate more accurately (lower DIFF, higher MP).
We further evaluate the multi-head mechanism of DCBlock by changing the number of text-adaptive convolution kernels to 1, 4, 16 (DC1, DC4, DC16).
The results show that more convolution kernels do not mean better performance.
The reason may be that more convolution kernels reduce the convergence efficiency of the DE-Net.
Therefore, we set 8 text-adaptive convolution kernels in DCBlock.

\noindent \textbf{Effect of Semantic Decoder (SD). } Without the Semantic Decoder (B w/ DE w/ CP w/DC), the model predicts the spatial editing parameters from the feature provided by the source image encoder. 
Armed with the Semantic Decoder, the model (B w/ DE w/ CP w/DC w/SD ) improves the IS, SIM, MP, and decreases the DIFF. 
The results prove that the semantic decoder can promote the DCBlock to make more effective and accurate manipulations.

\subsection{Visualisation results of DCBlock}
\begin{figure}[t] \small
  \centering
  \includegraphics[width=\linewidth]{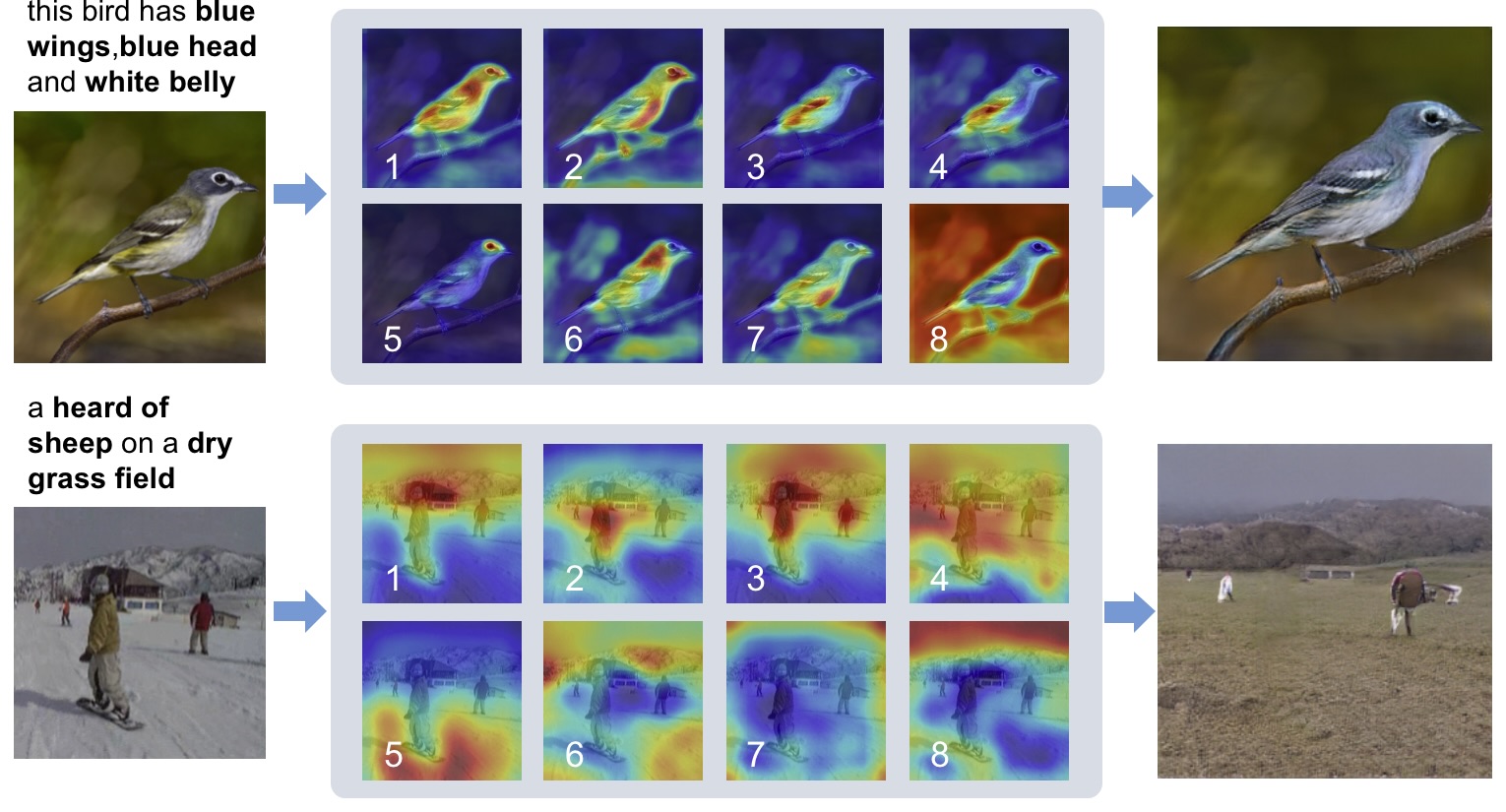}
  \caption{Visualization of the queried results by different text-adaptive convolution kernels.}
  \label{fig6}
\end{figure}

To further demonstrate the effectiveness of DCBlock, we visualize the text-conditioned queried results at the last DCBlock in the target image decoder.
As shown in Figure~\ref{fig6}, the kernel attends to different semantics in source image features.
In addition, we find that the convolution kernel can be divided into two types, one focusing on text-related image parts and the other focusing on text-irrelevant image parts.
For the CUB example, kernel 8 segments the whole background, which is text-irrelevant image parts.
The other kernel attends on text-related image parts, e.g., ``wings'', ``head'', ``belly''.
For the COCO example, kernel 8 segments the sky, which is text-irrelevant image parts.
The other kernel attends to text-related image parts, e.g., ground, human, and mountain, which need to be edited.
It demonstrates that our DCBlock can attend to different semantic parts and distinguish between text-related and text-irrelevant parts.


\section{Conclusion}

\vspace{-0.3cm}

In this paper, we propose a novel DE-Net for the text-guided image editing task. 
Compared with previous models, our DE-Net can manipulate the source images according to text guidance more correctly and accurately.
Via DE-Net, we propose a Dynamic text-guided Editing Block (DEBlock) to enable our model can deal with a variety of editing tasks adaptively through the dynamic composition of different editing modules.
We also propose a novel Composition Predictor (Comp-Pred) that compares the source image features and given text guidance, and predict the combination weights for DEBlock.
Furthermore, we propose a new Dynamic text-adaptive Convolution Block (DCBlock) with a semantic decoder to help the target image decoder distinguish between text-required and text-irrelevant parts.
Extensive experiment results demonstrate that the proposed DE-Net significantly outperforms state-of-the-art models on both CUB and COCO datasets.

{\small
\bibliographystyle{ieee_fullname}
\bibliography{egbib}
}

\end{document}